\documentclass[reprint,aps,amsmath,amssymb,prl,floatfix,superscriptaddress]{revtex4-2}
\usepackage[english]{babel}

\usepackage{subfigure,graphicx,float}
\usepackage{amsfonts,amsmath,amssymb,amsthm}
\usepackage[colorlinks,allcolors =blue]{hyperref}
\usepackage[all]{hypcap}

\begin{document}
\title{Parallel Machine Learning for Forecasting the Dynamics of Complex Networks}

\author{Keshav Srinivasan}
\affiliation{University of Maryland, College Park, Maryland 20742}

\author{Nolan Coble}
\affiliation{University of Maryland, College Park, Maryland 20742}
\affiliation{SUNY Brockport, New York 14420, USA.}

\author{Joy Hamlin}
\affiliation{Stony Brook University, New York 11794, USA}

\author{Thomas Antonsen}
\affiliation{University of Maryland, College Park, Maryland 20742}

\author{Edward Ott}
\affiliation{University of Maryland, College Park, Maryland 20742}

\author{Michelle Girvan}
\affiliation{University of Maryland, College Park, Maryland 20742}

\date{\today}
\begin{abstract}
Forecasting the dynamics of large complex networks from previous time-series data is important in a wide range of contexts. Here we present a machine learning scheme for this task using a parallel architecture that mimics the topology of the network of interest. We demonstrate the utility and scalability of our method implemented using reservoir computing on a chaotic network of oscillators. Two levels of prior knowledge are considered: (i) the network links are known; and  (ii) the network links are unknown and inferred via a data-driven approach to approximately optimize prediction.
\end{abstract}

\maketitle

Machine learning (ML) has played a vital role in recent scientific advances in many disciplines. A key problem in these contexts is time series prediction of a dynamical system for which a first-principles, knowledge-based description is unavailable \cite{jaeger_2004}. By using ML in combination with measured time-series data, one can hope to construct a faithful model of a system's dynamics and to then use this model to predict the future evolution of the system's state. Our aim in this paper is to address this goal for \emph{large systems of interacting components with complex connectivity and dynamics} - a system type of enormous technical and scientific interest in many fields, ranging, e.g., from neuroscience to power grids. However, straightforward application of the standard ML prediction schemes becomes problematic when applied to forecasting the dynamics of large networks. To deal with such systems, we propose a parallel forecasting method for networks with complex dynamics. In our approach, we construct an ML architecture that mimics the topology of the network. Each node of the network to be predicted is assigned an individual small ML device and these individual ML devices are linked to each other based on the underlying connectivity of the network (either known \emph{a priori} or inferred from the available time series data). We demonstrate and test this method by applying it to a network of Kuramoto oscillators \cite{kuramoto_1975,acebron_2005} constructed to exhibit chaotic dynamics. Our method is motivated in part by previous work on parallel ML prediction of large spatiotemporally chaotic systems \cite{pathak_2018,arcomano_2020}.

We consider two scenarios: (a) the connectivity of the oscillator network is known, and (b) the connectivity of the oscillator network is unknown \emph{a priori}, yet may be approximately inferred from node time series data. Scenario (a) serves two purposes: first, as preparation for the more challenging situation presented by scenario (b), and second, as a method applicable to cases where the connectivity is, in fact, known. The main conclusion of our paper is that our proposed parallel ML scheme enables data-based network dynamics prediction in cases that would otherwise (i.e., without parallelization) be unattainable.

In order to demonstrate and test our approach, we consider the well-studied Kuramoto model of $N$ network-coupled oscillators,
\begin{equation}
\dot{\theta_i} = \omega_i + K\sum_{j=1}^{N}A_{ij}\sin(\theta_j - \theta_i),
\label{eq:kuramoto}
\end{equation} 
where $\theta_i$ is the phase angle of oscillator $i$, $\omega_i$ is the natural frequency of oscillator $i$  when uncoupled, $K$ is the coupling strength, and ${A_{ij}}$ is the adjacency matrix that specifies the structure of the oscillator network ($A_{ij} = 1$, if there exists a network link from node $j$ to node $i$ with $i\neq j$, and $A_{ij} = 0$, otherwise). Here we consider an undirected ($A_{ij} = A_{ji}$), frequency assortative Kuramoto network \cite{restrepo_2014}. By `frequency assortative' we mean that two nodes are more likely to be linked if their natural oscillation frequencies are numerically close. The resulting frequency assortative system has chaotic dynamics for certain choices of parameters \cite{skardal_2015}, hence serving as a good example of complex network dynamics. Each node is taken to have the same number of connections (this number is called the node's degree). The oscillator natural frequencies, $\omega_i$, are drawn from a uniform random distribution from $-\pi/2$ to $\pi/2$. The frequency assortative network (i.e., the set of matrix elements $A_{ij}$) is constructed by starting with $N_o$ unlinked nodes, each with its assigned frequency ($\omega_{i}$ for node $i$) and then successively adding links, as follows. After randomly choosing a node $i$ that still requires additional links, we next randomly pick another node $j$ (not already connected to node $i$) which also still requires additional links, and then, with probability $p_{ij}$, we link nodes $i$ and $j$, where
\begin{equation}
p_{ij} \propto \frac{\delta^\gamma}{\delta^\gamma + |\omega_i-\omega_j|^\gamma}
\label{eq:assortivity}
\end{equation} 
We continue in this way to make links until all nodes have the desired degree. Due to the form of $p_{ij}$ (Eq. (\ref{eq:assortivity})) nodes with similar natural frequencies are connected with a higher probability (see Supplementary Fig. 1).
We use the global order parameter, $R$, as a metric to measure the dynamics of the oscillator network, where
\begin{equation}
R(t)=\sum_{i=1}^{N} \sum_{j=1}^{N} A_{i j} e^{i \theta_{j}}
\end{equation} 
\begin{figure}[H]
\centering
\includegraphics[width=.45\textwidth]{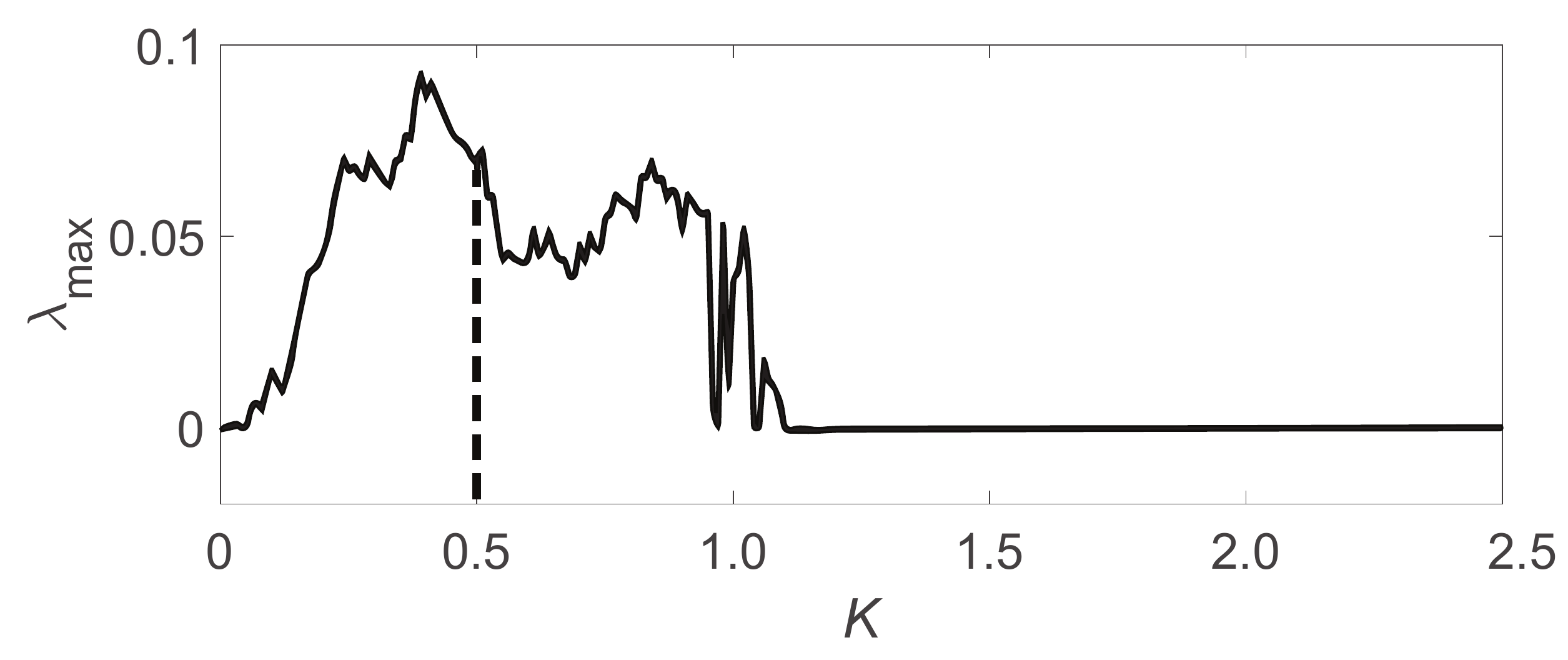}
\hfill
\caption{Largest Lyapunov Exponent as a function of the coupling constant, $K$. The dashed line represents the chosen value of $K$ for our studies.}
\label{fig:LLE}
\end{figure}
\noindent

For our frequency assortative network, with $N=50$, a nodal degree of 3, $\delta=0.8$, $\gamma=5$ and $K=0.5$ (our standard parameter set for most of out subsequent numerical experiments), we observe chaotic behavior, which is confirmed by the positive value of $\lambda_{max}$, the largest Lyapunov exponent of the system (Fig. \ref{fig:LLE}).\\

\textit{Background on non-parallel Reservoir Computing prediction}. In this paper we use Reservoir Computing (RC) \cite{jaeger_2001,maass_2002} as our ML scheme, because of its demonstrated utility for time series prediction \cite{jaeger_2004,dambre_2012,canaday_2018}. We consider a reservoir computer constructed with an artificial high dimensional dynamical system, known as the reservoir, which is coupled to an input through an input layer, specified by a matrix $\mathbf{W_{in}}$ which maps the input vector, $\mathbf{u}$, at discrete time $t$, to the reservoir state variables, which are collectively expressed as the scalar components of the reservoir state vector $\mathbf{r}$. In our RC implementation, the reservoir is a network (not to be confused with the network, e.g., Eq. (\ref{eq:kuramoto}), whose state we desire to predict), and the $k$th component of the vector $\mathbf{r}$ is the scalar state of reservoir node $k$. The RC network is directed, sparse, and random  with $N_r$ nodes having average input degree, $\kappa=3$. The RC adjacency matrix is denoted $B$, with matrix elements $B_{kk}=0$, and $B_{kl}$ for $k\neq l$ chosen randomly and uniformly from [$-\beta$, $\beta$] where $\beta$ is chosen to yield a maximum eigenvalue of $B$ denoted $\rho$ (known as the spectral radius). Each input to the reservoir is sent to $N_r/N_{in}$ reservoir nodes, where $N_{in}$ is the number of inputs to the RC (Note: $N_r$ is chosen to be an integer multiple of $N_{in}$). The input matrix, $\mathbf{W_{in}}$, is then a $N_r\times N_{in}$ dimensional matrix. The elements of $\mathbf{W_{in}}$ are chosen so that every node in the reservoir receives exactly one input from $\mathbf{u}(t)$ while each input in $\mathbf{u}(t)$ is connected to $N_r/N_{in}$ nodes in the reservoir network (see Supplementary Material for further discussion). The non-zero elements are drawn from a uniform random distribution from [$-\sigma$,$\sigma$], where $\sigma$ is the input scaling. The reservoir state, $\mathbf{r}(t)$, is taken to evolve according to 
\begin{equation}
    \mathbf{r}(t+\Delta t) = \alpha \mathbf{r}(t)+(1-\alpha)\tanh [\mathbf{Br}(t)+ \mathbf{W_{in}}\mathbf{u}(t)],
    \label{eq:rc1}
\end{equation} 
where the tanh function is applied component-wise to its vector argument. Here $\alpha$ is the leak rate which controls the timescale of the reservoir nodes. The output of the system, $ \Tilde{\mathbf{u}}, $ is defined through the output layer and is given by
\begin{equation}
    \Tilde{\mathbf{u}}(t) = \mathbf{W_{out}}\mathbf{r}(t).
    \label{eq:rc_out}
\end{equation} 
For the task of time-series prediction, the reservoir computer is used in two different modes: a training mode and a prediction mode. In the training mode, the reservoir computing system, represented by Eqs. (\ref{eq:rc1}) and (\ref{eq:rc_out}), is run for the time interval over which training data $u(t)=u(n\Delta t)\ (n=-n_t,(1-n_t),(2-n_t),...,0)$ is available, $\mathbf{r}(n\Delta t)$ is computed, and the output matrix $\mathbf{W_{out}}$ is adjusted (`trained') so that the output of the reservoir computer $\Tilde{\mathbf{u}}(t)$ best approximates $\mathbf{u}(t)$. This is done through a ridge regression procedure, wherein we minimize the error summed over the training times $t=n\Delta t$ for $n$ running from $1-n_t$ to $0$, 
\begin{equation}
    \label{eq:Tikhonov}
    \min _{\mathbf{W}_{\text {out}}}\left\{\sum\left[\left\|\mathbf{W}_{\text {out}} \mathbf{r}(t)-\mathbf{u}(t)\right\|^{2}\right]+\beta \operatorname{Tr}\left(\mathbf{W}_{\text {out}} \mathbf{W}_{\text {out}}^{T}\right)\right\}
\end{equation} 
Here $\beta$ is the Tikhonov regularization parameter that is used to prevent over-fitting. The quantities ($N_r,\ \rho ,\ \sigma ,\ \alpha \ \text{and }\beta$), referred to as hyperparameters of the reservoir computing setup, are collectively used to control the performance of system. In this paper we chose the hyperparameters by a subsequent iterative process approximately maximizing the valid prediction time (See Eq. (\ref{eq:error})) over the hyperparamters via a coarse grid search (See Supplementary Material Section III). In the prediction mode, the reservoir state now evolves autonomously in ``closed-loop" mode; i.e., the output at time $t$, now serves as the input at time $t+\Delta t$,
\begin{equation}
    \mathbf{r}(t+\Delta t) = \tanh [\mathbf{Br}(t)+ \mathbf{W_{in}}\mathbf{W_{out}}\mathbf{r}(t)].
    \label{eq:training}
\end{equation} 
This procedure generates a predicted time series $\hat{\mathbf{u}} ( n\Delta t)=\mathbf{W_{out}}\mathbf{r}(t)$ that is assumed to approximate the true future evolution of the state of the system, $\mathbf{u}(t)$ at a time $n\Delta t$ for $n>0$ (we choose $\Delta t$ small compared to the time scale for variation of $u$ so that $u (n\Delta t)$ essentially specifies the continuous time function $\mathbf u (t)$).\\

\begin{figure}[htp]
\centering
\includegraphics[width=.5\textwidth]{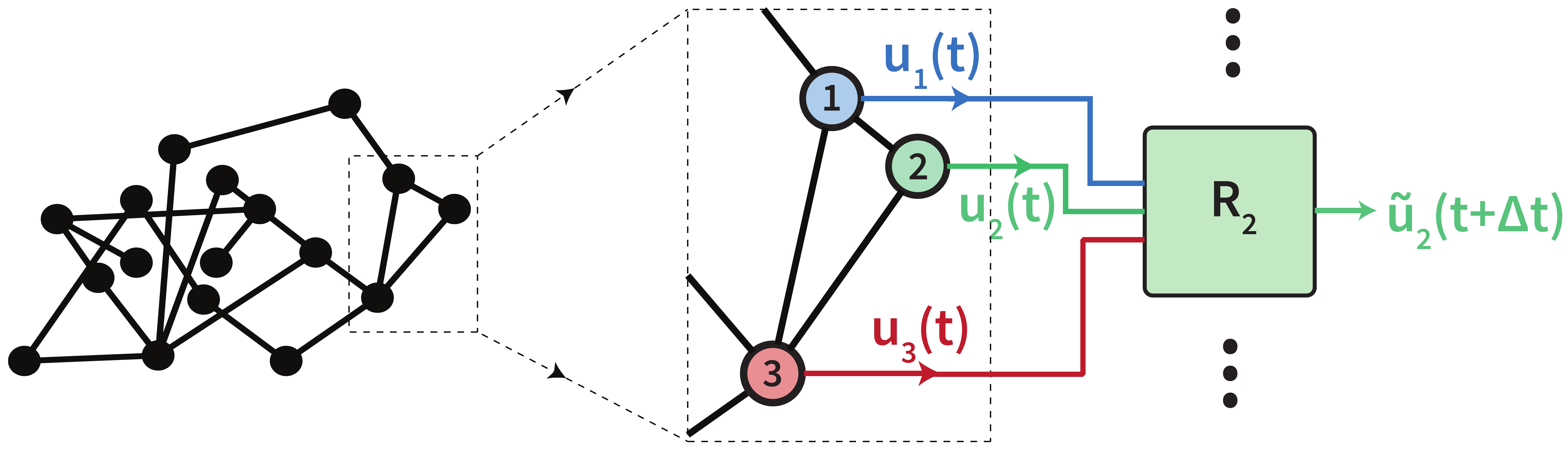}
\caption{A schematic diagram for the parallel network ML architecture. Here we show Reservoir 2 (R$_2$), which receives input from its assigned node (node 2), plus inputs from nodes connected to node 2 (i.e., nodes 1 and 3). R$_2$ is then is then trained to predict its assigned node (node 2). This process is the same for each node in the network, such that the connectivity among the reservoirs mimics the network to be predicted.}
\label{fig:schematic}
\end{figure}

\textit{Parallel ML scheme for network prediction.} In order to address the high computational complexity of predicting large networks, we introduce a parallel network RC architecture (see the schematic in Fig. \ref{fig:schematic}). Each node, $i$, in the predicted network is assigned its own reservoir, $R_i$. The inputs to this reservoir are the signal of node $i$ itself, as well as that of the nearest network neighbors of node $i$. The number of such neighbors is equal to the network degree. The reservoir $R_i$ is then trained on these inputs to predict the signal of node $i$. Because each $R_i$ predicts just one node, its size $N_r$ can be relatively small. In addition, since our parallel scheme uses an interconnected network of independently trained reservoirs, we can efficiently parallelize our training process, making the system scalable to large networks.\\

\textit{Results.} To compare the parallel, multiple RC scheme with the single RC approach, we use a $N_o=50$ node frequency assortative Kuramoto oscillator ($\delta=0.8$. $\gamma=5$) network with a coupling constant of $K=0.5$. We study the magnitude of the global order parameter $|R|$ which tells us about network-level activity (see Fig. 3) and the prediction of the evolution of individual node states (see Supplementary Figure 3), both of which show the same main qualitative behavior. For the purpose of forming inputs to the reservoir, we specify the state of the oscillator $i$ as [$\sin \theta_i(t),\cos\theta_i(t)$]. The input matrix is generated as described above and in Supplementary Material, Section 2. \\

\begin{figure}
\includegraphics[width=.5\textwidth]{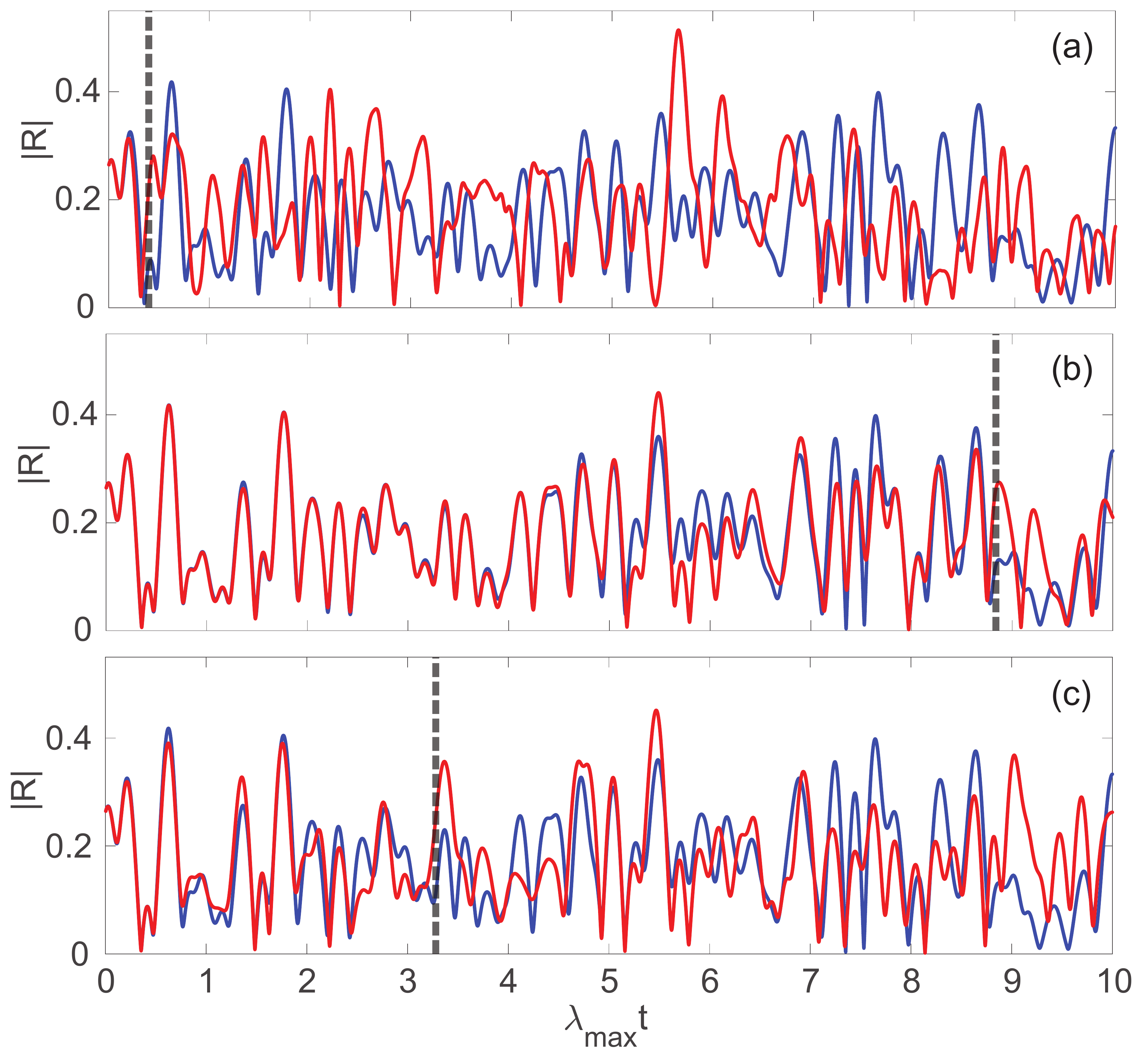}

\caption{Prediction of the order parameter in 3 different cases of ML implementations. The blue curves are the data (the truth) and red curves are predictions. The dotted lines in each plot denote the valid prediction time. (a) Single, non-parallel RC prediction using a large reservoir ($N_r=10000$), (b) Parallel prediction with known network links, using 50 separate reservoirs each having modest size ($N_r=200$), (c) Parallel prediction with unknown network links, using 50 separate reservoirs each having modest size ($N_r=200$). The network structure is estimated by using transfer entropy as a metric to draw network edges.}
\label{fig:predictions}
\end{figure}

\textit{Single non-parallel reservoir prediction}. The single reservoir computer prediction can fail as the size of the network we want to forecast increases. This is clearly demonstrated in Fig. \hyperref[fig:predictions]{3(a)}, where the prediction breaks down in a fraction of a Lyapunov time, $\lambda_{max}t$. We quantify the duration of an accurate prediction by a metric that we call the “valid prediction time”. This metric is defined as the amount of time elapsed before the normalized root mean squared prediction error (NRMSE), $E(t)$, exceeds some chosen value $f$, $0 < f < 1$, for the the first time, where
\begin{equation}
    E(t)=\frac{\|\mathbf{u}(t)-\widetilde{\mathbf{u}}(t)\|}{\left\langle\|\mathbf{u}(t)\|^{2}\right\rangle^{1 / 2}}.
    \label{eq:error}
\end{equation} 
The valid prediction time for $f=0.1$ is marked in Fig. \ref{fig:predictions} by a vertical dotted lines. Even for the very large reservoir ($N_{r}$=10000), close to the limit of our computer resources, that is used in Fig. \hyperref[fig:predictions]{3(a)}, the system is still not able to predict past a fraction of a Lyapunov time.\\

\textit{Parallel scheme with known links}. In cases where the network structure is known \textit{a priori}, such as in certain social networks, and we can construct our parallel reservoir architecture by using the known network links. In the case of our Kuramoto oscillator network, we demonstrate our results on the 50 node network by using 50 separate parallel reservoirs of relatively modest size ($N_r = 200$ as compared to $N_r = 10000$ for Fig. \hyperref[fig:predictions]{3(a)}), each having the same set of hyperparameters [see Fig. \hyperref[fig:predictions]{3(b)}]. The predictive performance of this architecture could potentially be enhanced by individually optimizing hyperparameters for each of the 50 reservoirs, but this would considerably increase both the time and computational resources required for this task. As seen from Fig. \hyperref[fig:predictions]{3(b)}, our parallel scheme does exceedingly well for multiples of the Lyapunov time, $\lambda_{max}t$. This is particularly clear from a comparison of valid prediction time (vertical dashed lines) in Fig. \hyperref[fig:predictions]{3(a)} versus those in Fig. \hyperref[fig:predictions]{3(b)}, the latter being $\gtrsim 10$ times larger, while at the same time being much less computationally demanding (mainly due to the difference in $N_r$, $N_r = 10000$ for the nonparallel case versus $N_r = 200$ in the parallel case).\\

\textit{Parallel scheme with unknown links}. In many cases of interest, one may not have information about the underlying network structure. Using nodal time-series data for finding links in networks, such as metabolic \cite{holme_2005} and gene-regulatory networks \cite{banf_2017}, is an active area of current research. Many heuristic-based \cite{lu_2011} and statistics-based tools like conditional mutual information \cite{tan_2014}, and correlation \cite{kumar_2012}, as well as a machine learning technique \cite{banerjee_2019}, have been used for link inference and might give useful approximations of the underlying network structure. These methods could then potentially be used in our parallel network scheme. As an example, we now demonstrate the performance of our parallel method combined with the use of Transfer Entropy \cite{schreiber_2000} for link inference. Transfer entropy is a statistical method to infer causal relationships between variables by using conditional probabilities: If a signal A has a causal effect on signal B, then the probability of B conditioned on its past is different from the probability of B conditioned on both its past and the past of A. Transfer entropy can also be expressed in terms of the conditional mutual information as
\begin{equation}
    T_{X\xrightarrow{}Y}=I(Y_t;X_{t-1:t-L}|Y_{t-1:t-L}).
    \label{eq:Transfer_Entropy}
\end{equation} 
Considering the problem of network state prediction, we use past measured nodal state time-series to calculate the transfer entropy between each pair of nodes in the network using Eq. (\ref{eq:Transfer_Entropy}) and then pick a threshold. Pairs of nodes with transfer entropy values above the threshold are assigned a link between them. As we decrease this threshold, we draw more links and hence increase the average number of supposed neighbors for each node. Initially, decreasing the threshold, or in other words increasing the number of supposed neighbors, increases the number of true positive links and improves the predictive performance of our reservoir scheme. But if this threshold is decreased too much, the number of false positives increases drastically degrading the predictions. Since our goal is prediction, we view the link-inference threshold on the transfer entropy as an additional hyperparameter and chose it (along with the other hyperparameters), so as to optimize the valid prediction time. By this procedure, we effectively bootstrap our prediction process to determine the link threshold criterion. An example set of results for $N_r \approx 200$ (See Supplementary Material Section II for details) is shown in Fig. \hyperref[fig:predictions]{3(c)}. Again, in marked contrast with the results in Fig. \hyperref[fig:predictions]{3(a)} for a large single RC ($N_r=10000$), we obtain good predictions, e.g., a valid time between 3 and 4 Lyapunov times for $|R|$.\\

\begin{figure}
\centering
\includegraphics[width=0.5\textwidth]{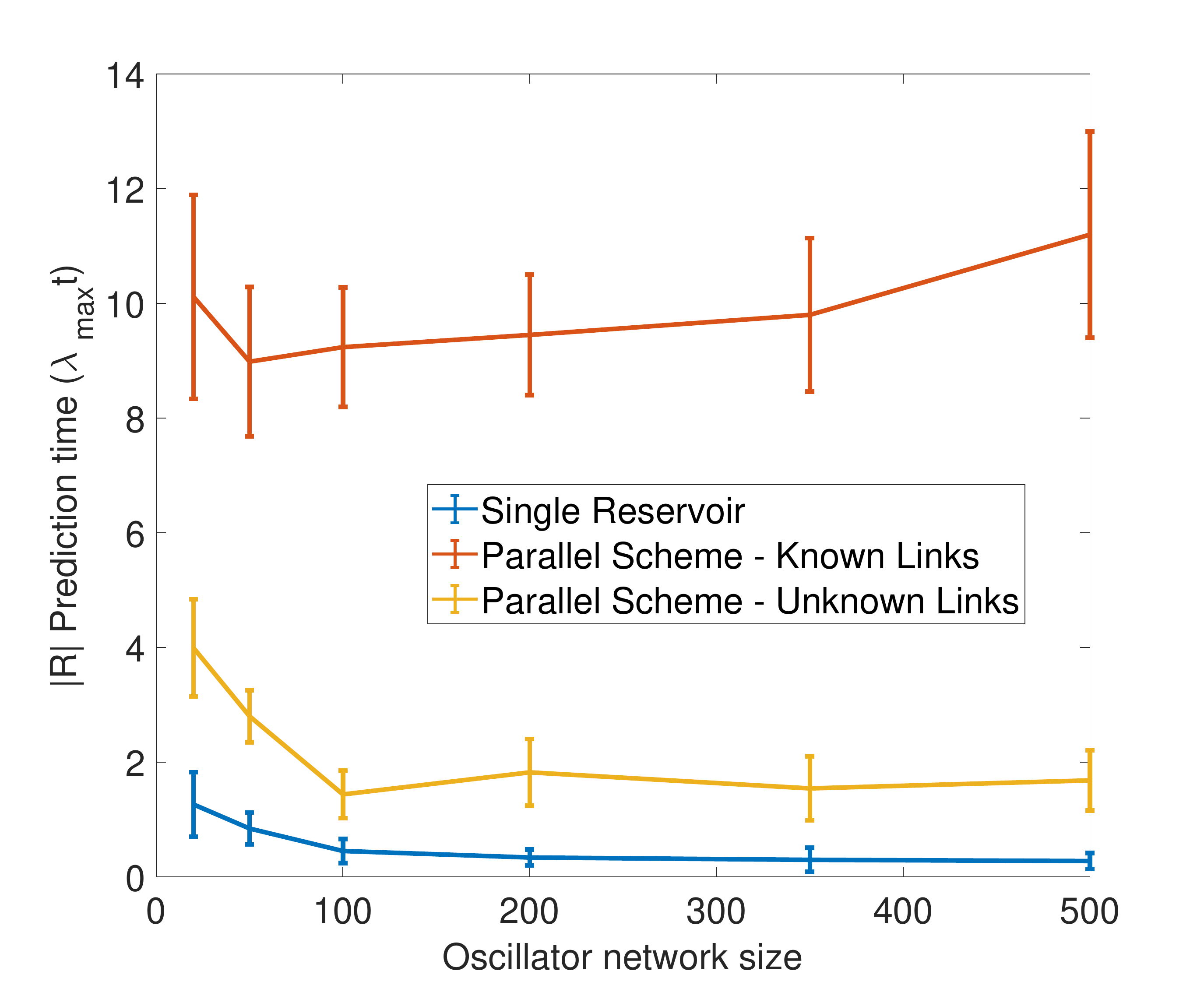}

\caption{Performance of the different Reservoir Computing methods as a function of the Kuramoto oscillator network size.\label{fig:scaling}}
\end{figure}

\textit{Dependence on the size of the predicted network}. Fig. ~\ref{fig:scaling} shows a plot of the valid prediction time as a function of the oscillator network size, $N_o$. As we increase the size of the oscillator network, the prediction using a single reservoir ($N_r=10000$) quickly degrades even further and becomes unable to capture the network dynamics at all. Since the parallel method assigns a reservoir to each oscillator in the network, for the case with known links, as expected it maintains constant performance to within the estimated uncertainty of the valid times. However, when we do not know the links, the ability of our parallel scheme to predict the network dynamics is limited by its ability to make accurate link predictions. The reduction in performance incurred by missing real links (i.e. false negatives) far outweighs that associated with false positive links. As we increase the oscillator network size, the accuracy of link determination gets worse which ultimately affects the predictive performance of this method. For small oscillator network size of $N_o=10$, our true positive rate is 100\%, while our false discovery rate is 37.5\%, but as the network science increases to $N_o=500$, our true positive rate drops to 96\% and our false discovery rate becomes 69.6\%. That the prediction degradation is more sensitive to a false negative link inference than to a false positive link inference can be understood as follows. The false negative inference of a link to node $i$ deprives reservoir $R_i$ of vital information needed for prediction of the state of node $i$. In contrast, reservoir $R_i$ can compensate for a false positive link from node $j$ to node $i$ by learning, through its training, to ignore its time series input from node $j$. However, if there are too many false positive links to node $i$, reservoir $R_i$ becomes overburdened, and its state prediction accuracy degrades.\\

\textit{Conclusion.} We are able to construct accurate, data-driven forecasts for the dynamics of large complex networks using a parallel ML architecture that reflects the topology of the network to be predicted. In cases for which a non-parallel approach with comparable resources fails, our scheme is successful when the network links are either known or unknown {\emph a priori}. The parallel nature makes our approach scalable for extremely large networks, creating potential applications to many fields. \\
    
This work was supported by the National Science Foundation under Grant Nos. PHY-1461089, DGE-1632976, and DMS-1813027. 

\bibliography{citations}

\begin{thebibliography}{18}%
\makeatletter
\providecommand \@ifxundefined [1]{%
 \@ifx{#1\undefined}
}%
\providecommand \@ifnum [1]{%
 \ifnum #1\expandafter \@firstoftwo
 \else \expandafter \@secondoftwo
 \fi
}%
\providecommand \@ifx [1]{%
 \ifx #1\expandafter \@firstoftwo
 \else \expandafter \@secondoftwo
 \fi
}%
\providecommand \natexlab [1]{#1}%
\providecommand \enquote  [1]{``#1''}%
\providecommand \bibnamefont  [1]{#1}%
\providecommand \bibfnamefont [1]{#1}%
\providecommand \citenamefont [1]{#1}%
\providecommand \href@noop [0]{\@secondoftwo}%
\providecommand \href [0]{\begingroup \@sanitize@url \@href}%
\providecommand \@href[1]{\@@startlink{#1}\@@href}%
\providecommand \@@href[1]{\endgroup#1\@@endlink}%
\providecommand \@sanitize@url [0]{\catcode `\\12\catcode `\$12\catcode
  `\&12\catcode `\#12\catcode `\^12\catcode `\_12\catcode `\%12\relax}%
\providecommand \@@startlink[1]{}%
\providecommand \@@endlink[0]{}%
\providecommand \url  [0]{\begingroup\@sanitize@url \@url }%
\providecommand \@url [1]{\endgroup\@href {#1}{\urlprefix }}%
\providecommand \urlprefix  [0]{URL }%
\providecommand \Eprint [0]{\href }%
\providecommand \doibase [0]{https://doi.org/}%
\providecommand \selectlanguage [0]{\@gobble}%
\providecommand \bibinfo  [0]{\@secondoftwo}%
\providecommand \bibfield  [0]{\@secondoftwo}%
\providecommand \translation [1]{[#1]}%
\providecommand \BibitemOpen [0]{}%
\providecommand \bibitemStop [0]{}%
\providecommand \bibitemNoStop [0]{.\EOS\space}%
\providecommand \EOS [0]{\spacefactor3000\relax}%
\providecommand \BibitemShut  [1]{\csname bibitem#1\endcsname}%
\let\auto@bib@innerbib\@empty
\bibitem [{\citenamefont {Jaeger}\ and\ \citenamefont
  {Haas}(2004)}]{jaeger_2004}%
  \BibitemOpen
  \bibfield  {author} {\bibinfo {author} {\bibfnamefont {H.}~\bibnamefont
  {Jaeger}}\ and\ \bibinfo {author} {\bibfnamefont {H.}~\bibnamefont {Haas}},\
  }\href {https://doi.org/10.1126/science.1091277} {\bibfield  {journal}
  {\bibinfo  {journal} {Science}\ }\textbf {\bibinfo {volume} {304}},\ \bibinfo
  {pages} {78} (\bibinfo {year} {2004})}\BibitemShut {NoStop}%
\bibitem [{\citenamefont {Kuramoto}(1975)}]{kuramoto_1975}%
  \BibitemOpen
  \bibfield  {author} {\bibinfo {author} {\bibfnamefont {Y.}~\bibnamefont
  {Kuramoto}},\ }in\ \href {https://doi.org/10.1007/BFb0013365} {\emph
  {\bibinfo {booktitle} {International {Symposium} on {Mathematical} {Problems}
  in {Theoretical} {Physics}}}},\ \bibinfo {series and number} {Lecture {Notes}
  in {Physics}}\ (\bibinfo  {publisher} {Springer},\ \bibinfo {address}
  {Berlin, Heidelberg},\ \bibinfo {year} {1975})\ pp.\ \bibinfo {pages}
  {420--422}\BibitemShut {NoStop}%
\bibitem [{\citenamefont {Acebrón}\ \emph {et~al.}(2005)\citenamefont
  {Acebrón}, \citenamefont {Bonilla}, \citenamefont {Pérez~Vicente},
  \citenamefont {Ritort},\ and\ \citenamefont {Spigler}}]{acebron_2005}%
  \BibitemOpen
  \bibfield  {author} {\bibinfo {author} {\bibfnamefont {J.~A.}\ \bibnamefont
  {Acebrón}}, \bibinfo {author} {\bibfnamefont {L.~L.}\ \bibnamefont
  {Bonilla}}, \bibinfo {author} {\bibfnamefont {C.~J.}\ \bibnamefont
  {Pérez~Vicente}}, \bibinfo {author} {\bibfnamefont {F.}~\bibnamefont
  {Ritort}},\ and\ \bibinfo {author} {\bibfnamefont {R.}~\bibnamefont
  {Spigler}},\ }\href {https://doi.org/10.1103/RevModPhys.77.137} {\bibfield
  {journal} {\bibinfo  {journal} {Reviews of Modern Physics}\ }\textbf
  {\bibinfo {volume} {77}},\ \bibinfo {pages} {137} (\bibinfo {year}
  {2005})}\BibitemShut {NoStop}%
\bibitem [{\citenamefont {Pathak}\ \emph {et~al.}(2018)\citenamefont {Pathak},
  \citenamefont {Hunt}, \citenamefont {Girvan}, \citenamefont {Lu},\ and\
  \citenamefont {Ott}}]{pathak_2018}%
  \BibitemOpen
  \bibfield  {author} {\bibinfo {author} {\bibfnamefont {J.}~\bibnamefont
  {Pathak}}, \bibinfo {author} {\bibfnamefont {B.}~\bibnamefont {Hunt}},
  \bibinfo {author} {\bibfnamefont {M.}~\bibnamefont {Girvan}}, \bibinfo
  {author} {\bibfnamefont {Z.}~\bibnamefont {Lu}},\ and\ \bibinfo {author}
  {\bibfnamefont {E.}~\bibnamefont {Ott}},\ }\href
  {https://doi.org/10.1103/PhysRevLett.120.024102} {\bibfield  {journal}
  {\bibinfo  {journal} {Physical Review Letters}\ }\textbf {\bibinfo {volume}
  {120}},\ \bibinfo {pages} {024102} (\bibinfo {year} {2018})}\BibitemShut
  {NoStop}%
\bibitem [{\citenamefont {Arcomano}\ \emph {et~al.}(2020)\citenamefont
  {Arcomano}, \citenamefont {Szunyogh}, \citenamefont {Pathak}, \citenamefont
  {Wikner}, \citenamefont {Hunt},\ and\ \citenamefont {Ott}}]{arcomano_2020}%
  \BibitemOpen
  \bibfield  {author} {\bibinfo {author} {\bibfnamefont {T.}~\bibnamefont
  {Arcomano}}, \bibinfo {author} {\bibfnamefont {I.}~\bibnamefont {Szunyogh}},
  \bibinfo {author} {\bibfnamefont {J.}~\bibnamefont {Pathak}}, \bibinfo
  {author} {\bibfnamefont {A.}~\bibnamefont {Wikner}}, \bibinfo {author}
  {\bibfnamefont {B.~R.}\ \bibnamefont {Hunt}},\ and\ \bibinfo {author}
  {\bibfnamefont {E.}~\bibnamefont {Ott}},\ }\bibfield  {journal} {\bibinfo
  {journal} {Geophysical Research Letters}\ }\textbf {\bibinfo {volume} {47}},\
  \href {https://doi.org/10.1029/2020GL087776} {10.1029/2020GL087776} (\bibinfo
  {year} {2020})\BibitemShut {NoStop}%
\bibitem [{\citenamefont {Restrepo}\ and\ \citenamefont
  {Ott}(2014)}]{restrepo_2014}%
  \BibitemOpen
  \bibfield  {author} {\bibinfo {author} {\bibfnamefont {J.~G.}\ \bibnamefont
  {Restrepo}}\ and\ \bibinfo {author} {\bibfnamefont {E.}~\bibnamefont {Ott}},\
  }\href {https://doi.org/10.1209/0295-5075/107/60006} {\bibfield  {journal}
  {\bibinfo  {journal} {EPL (Europhysics Letters)}\ }\textbf {\bibinfo {volume}
  {107}},\ \bibinfo {pages} {60006} (\bibinfo {year} {2014})}\BibitemShut
  {NoStop}%
\bibitem [{\citenamefont {Skardal}\ \emph {et~al.}(2015)\citenamefont
  {Skardal}, \citenamefont {Restrepo},\ and\ \citenamefont
  {Ott}}]{skardal_2015}%
  \BibitemOpen
  \bibfield  {author} {\bibinfo {author} {\bibfnamefont {P.~S.}\ \bibnamefont
  {Skardal}}, \bibinfo {author} {\bibfnamefont {J.~G.}\ \bibnamefont
  {Restrepo}},\ and\ \bibinfo {author} {\bibfnamefont {E.}~\bibnamefont
  {Ott}},\ }\href {https://doi.org/10.1103/PhysRevE.91.060902} {\bibfield
  {journal} {\bibinfo  {journal} {Physical Review E}\ }\textbf {\bibinfo
  {volume} {91}},\ \bibinfo {pages} {060902} (\bibinfo {year}
  {2015})}\BibitemShut {NoStop}%
\bibitem [{\citenamefont {Jaeger}(2001)}]{jaeger_2001}%
  \BibitemOpen
  \bibfield  {author} {\bibinfo {author} {\bibfnamefont {H.}~\bibnamefont
  {Jaeger}},\ }\href@noop {} {\emph {\bibinfo {title} {The ‘echo state’
  approach to analyzing and training recurrent neural networks}}},\ \bibinfo
  {type} {GMD Report}\ \bibinfo {number} {148}\ (\bibinfo  {institution}
  {German National Research Center for Information Technology},\ \bibinfo
  {year} {2001})\BibitemShut {NoStop}%
\bibitem [{\citenamefont {Maass}\ \emph {et~al.}(2002)\citenamefont {Maass},
  \citenamefont {Natschläger},\ and\ \citenamefont {Markram}}]{maass_2002}%
  \BibitemOpen
  \bibfield  {author} {\bibinfo {author} {\bibfnamefont {W.}~\bibnamefont
  {Maass}}, \bibinfo {author} {\bibfnamefont {T.}~\bibnamefont
  {Natschläger}},\ and\ \bibinfo {author} {\bibfnamefont {H.}~\bibnamefont
  {Markram}},\ }\href {https://doi.org/10.1162/089976602760407955} {\bibfield
  {journal} {\bibinfo  {journal} {Neural Computation}\ }\textbf {\bibinfo
  {volume} {14}},\ \bibinfo {pages} {2531} (\bibinfo {year}
  {2002})}\BibitemShut {NoStop}%
\bibitem [{\citenamefont {Dambre}\ \emph {et~al.}(2012)\citenamefont {Dambre},
  \citenamefont {Verstraeten}, \citenamefont {Schrauwen},\ and\ \citenamefont
  {Massar}}]{dambre_2012}%
  \BibitemOpen
  \bibfield  {author} {\bibinfo {author} {\bibfnamefont {J.}~\bibnamefont
  {Dambre}}, \bibinfo {author} {\bibfnamefont {D.}~\bibnamefont {Verstraeten}},
  \bibinfo {author} {\bibfnamefont {B.}~\bibnamefont {Schrauwen}},\ and\
  \bibinfo {author} {\bibfnamefont {S.}~\bibnamefont {Massar}},\ }\href
  {https://doi.org/10.1038/srep00514} {\bibfield  {journal} {\bibinfo
  {journal} {Scientific Reports}\ }\textbf {\bibinfo {volume} {2}},\ \bibinfo
  {pages} {514} (\bibinfo {year} {2012})}\BibitemShut {NoStop}%
\bibitem [{\citenamefont {Canaday}\ \emph {et~al.}(2018)\citenamefont
  {Canaday}, \citenamefont {Griffith},\ and\ \citenamefont
  {Gauthier}}]{canaday_2018}%
  \BibitemOpen
  \bibfield  {author} {\bibinfo {author} {\bibfnamefont {D.}~\bibnamefont
  {Canaday}}, \bibinfo {author} {\bibfnamefont {A.}~\bibnamefont {Griffith}},\
  and\ \bibinfo {author} {\bibfnamefont {D.~J.}\ \bibnamefont {Gauthier}},\
  }\href {https://doi.org/10.1063/1.5048199} {\bibfield  {journal} {\bibinfo
  {journal} {Chaos: An Interdisciplinary Journal of Nonlinear Science}\
  }\textbf {\bibinfo {volume} {28}},\ \bibinfo {pages} {123119} (\bibinfo
  {year} {2018})}\BibitemShut {NoStop}%
\bibitem [{\citenamefont {Holme}\ and\ \citenamefont
  {Huss}(2005)}]{holme_2005}%
  \BibitemOpen
  \bibfield  {author} {\bibinfo {author} {\bibfnamefont {P.}~\bibnamefont
  {Holme}}\ and\ \bibinfo {author} {\bibfnamefont {M.}~\bibnamefont {Huss}},\
  }\href {https://doi.org/10.1098/rsif.2005.0046} {\bibfield  {journal}
  {\bibinfo  {journal} {Journal of the Royal Society Interface}\ }\textbf
  {\bibinfo {volume} {2}},\ \bibinfo {pages} {327} (\bibinfo {year}
  {2005})}\BibitemShut {NoStop}%
\bibitem [{\citenamefont {Banf}\ and\ \citenamefont {Rhee}(2017)}]{banf_2017}%
  \BibitemOpen
  \bibfield  {author} {\bibinfo {author} {\bibfnamefont {M.}~\bibnamefont
  {Banf}}\ and\ \bibinfo {author} {\bibfnamefont {S.~Y.}\ \bibnamefont
  {Rhee}},\ }\href {https://doi.org/10.1038/srep41174} {\bibfield  {journal}
  {\bibinfo  {journal} {Scientific Reports}\ }\textbf {\bibinfo {volume} {7}},\
  \bibinfo {pages} {41174} (\bibinfo {year} {2017})}\BibitemShut {NoStop}%
\bibitem [{\citenamefont {Lü}\ and\ \citenamefont {Zhou}(2011)}]{lu_2011}%
  \BibitemOpen
  \bibfield  {author} {\bibinfo {author} {\bibfnamefont {L.}~\bibnamefont
  {Lü}}\ and\ \bibinfo {author} {\bibfnamefont {T.}~\bibnamefont {Zhou}},\
  }\href {https://doi.org/10.1016/j.physa.2010.11.027} {\bibfield  {journal}
  {\bibinfo  {journal} {Physica A: Statistical Mechanics and its Applications}\
  }\textbf {\bibinfo {volume} {390}},\ \bibinfo {pages} {1150} (\bibinfo {year}
  {2011})}\BibitemShut {NoStop}%
\bibitem [{\citenamefont {Tan}\ \emph {et~al.}(2014)\citenamefont {Tan},
  \citenamefont {Xia},\ and\ \citenamefont {Zhu}}]{tan_2014}%
  \BibitemOpen
  \bibfield  {author} {\bibinfo {author} {\bibfnamefont {F.}~\bibnamefont
  {Tan}}, \bibinfo {author} {\bibfnamefont {Y.}~\bibnamefont {Xia}},\ and\
  \bibinfo {author} {\bibfnamefont {B.}~\bibnamefont {Zhu}},\ }\href
  {https://doi.org/10.1371/journal.pone.0107056} {\bibfield  {journal}
  {\bibinfo  {journal} {PLOS ONE}\ }\textbf {\bibinfo {volume} {9}},\ \bibinfo
  {pages} {e107056} (\bibinfo {year} {2014})}\BibitemShut {NoStop}%
\bibitem [{\citenamefont {Kumar}\ and\ \citenamefont {Deo}(2012)}]{kumar_2012}%
  \BibitemOpen
  \bibfield  {author} {\bibinfo {author} {\bibfnamefont {S.}~\bibnamefont
  {Kumar}}\ and\ \bibinfo {author} {\bibfnamefont {N.}~\bibnamefont {Deo}},\
  }\href {https://doi.org/10.1103/PhysRevE.86.026101} {\bibfield  {journal}
  {\bibinfo  {journal} {Physical Review E}\ }\textbf {\bibinfo {volume} {86}},\
  \bibinfo {pages} {026101} (\bibinfo {year} {2012})}\BibitemShut {NoStop}%
\bibitem [{\citenamefont {Banerjee}\ \emph {et~al.}(2019)\citenamefont
  {Banerjee}, \citenamefont {Pathak}, \citenamefont {Roy}, \citenamefont
  {Restrepo},\ and\ \citenamefont {Ott}}]{banerjee_2019}%
  \BibitemOpen
  \bibfield  {author} {\bibinfo {author} {\bibfnamefont {A.}~\bibnamefont
  {Banerjee}}, \bibinfo {author} {\bibfnamefont {J.}~\bibnamefont {Pathak}},
  \bibinfo {author} {\bibfnamefont {R.}~\bibnamefont {Roy}}, \bibinfo {author}
  {\bibfnamefont {J.~G.}\ \bibnamefont {Restrepo}},\ and\ \bibinfo {author}
  {\bibfnamefont {E.}~\bibnamefont {Ott}},\ }\href
  {https://doi.org/10.1063/1.5134845} {\bibfield  {journal} {\bibinfo
  {journal} {Chaos: An Interdisciplinary Journal of Nonlinear Science}\
  }\textbf {\bibinfo {volume} {29}},\ \bibinfo {pages} {121104} (\bibinfo
  {year} {2019})}\BibitemShut {NoStop}%
\bibitem [{\citenamefont {Schreiber}(2000)}]{schreiber_2000}%
  \BibitemOpen
  \bibfield  {author} {\bibinfo {author} {\bibfnamefont {T.}~\bibnamefont
  {Schreiber}},\ }\href {https://doi.org/10.1103/PhysRevLett.85.461} {\bibfield
   {journal} {\bibinfo  {journal} {Physical Review Letters}\ }\textbf {\bibinfo
  {volume} {85}},\ \bibinfo {pages} {461} (\bibinfo {year} {2000})}\BibitemShut
  {NoStop}%
\end{thebibliography}%
\end{document}


\title{Supplementary Material}

\author{Keshav Srinivasan}
\affiliation{University of Maryland, College Park, Maryland 20742, USA.}

\author{Nolan Coble}
\affiliation{University of Maryland, College Park, Maryland 20742, USA.}
\affiliation{SUNY Brockport, New York 14420, USA.}

\author{Joy Hamlin}
\affiliation{Stony Brook University, New York 11794, USA}

\author{Tom M. Antonsen}
\affiliation{University of Maryland, College Park, Maryland 20742, USA.}

\author{Edward Ott}
\affiliation{University of Maryland, College Park, Maryland 20742, USA.}

\author{Michelle Girvan}
\affiliation{University of Maryland, College Park, Maryland 20742, USA.}

\maketitle
\section{Frequency Assortative Kuramoto Oscillator Network}
\begin{figure}[h]
\centering
\includegraphics[width=.4\textwidth]{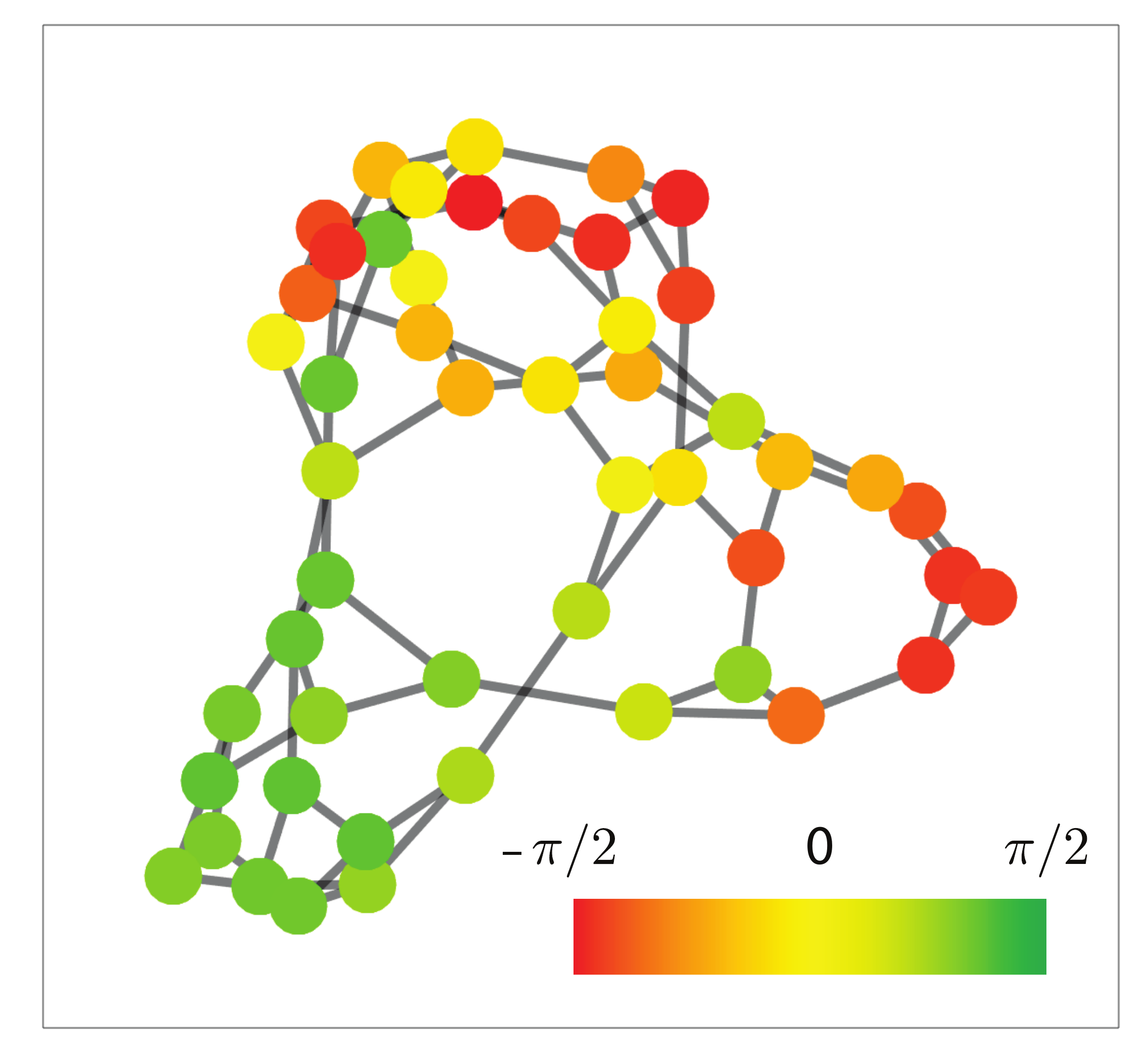}
\hfill
\caption{Frequency assoratative network structure with 50 nodes displaying the natural frequencies of the individual oscillators, with nodes colored by frequency. Similar frequencies are connected with higher probability.} 
\label{Suppfig:LLE}
\end{figure}

\section{Generating the Reservoir Input Matrix, $\mathbf{W_{in}}$}\label{SuppSec2}

In this section, we provide more details on how we generated the input matrix, $\mathbf{W_{in}}$, that feeds into the reservoir for predicting our Kuramoto oscillator network. Here the state of each oscillator, $i$, is specified as a tuple [$\sin \theta_i(t),\cos\theta_i(t)$]. The $N_{in}^{osc}$ input oscillators to a particular reservoir are collapsed into a single input vector, $\mathbf{u}$, with $N_{in}^{tot}$ ($N_{in}^{tot}=2N_{in}^{osc}$) elements  such that $\mathbf{u}_i(t)=\sin\theta_i(t),\ \mathbf{u}_{(N_{in}^{osc}+i)}(t)=\cos\theta_i(t) \text{ for } 1\leq i\leq N_{in}^{osc}$.  The elements of $\mathbf{W_{in}}$ are chosen so that each of the $N_r$ nodes in the reservoir receives exactly one input from amongst the $2N_{in}^{osc}$ components of the vector $\mathbf{u}(t)$. In the parallel scheme with known neighbors, all the $N_{in}^{tot}$ inputs to a reservoir are treated equally; i.e. each input in $\mathbf{u}(t)$ is connected to the same number ($N_r/N_{in}^{tot}$) of reservoir nodes (Note: $N_r$ is chosen to be an integer multiple of $N_{in}^{tot}$). The non-zero elements of $\mathbf{W_{in}}$ are then randomly drawn from a uniform distribution in [$-\sigma$,$\sigma$], where $\sigma$ is the input scaling.\\ 

\begin{figure}[h]
\centering
\includegraphics[width=.75\textwidth]{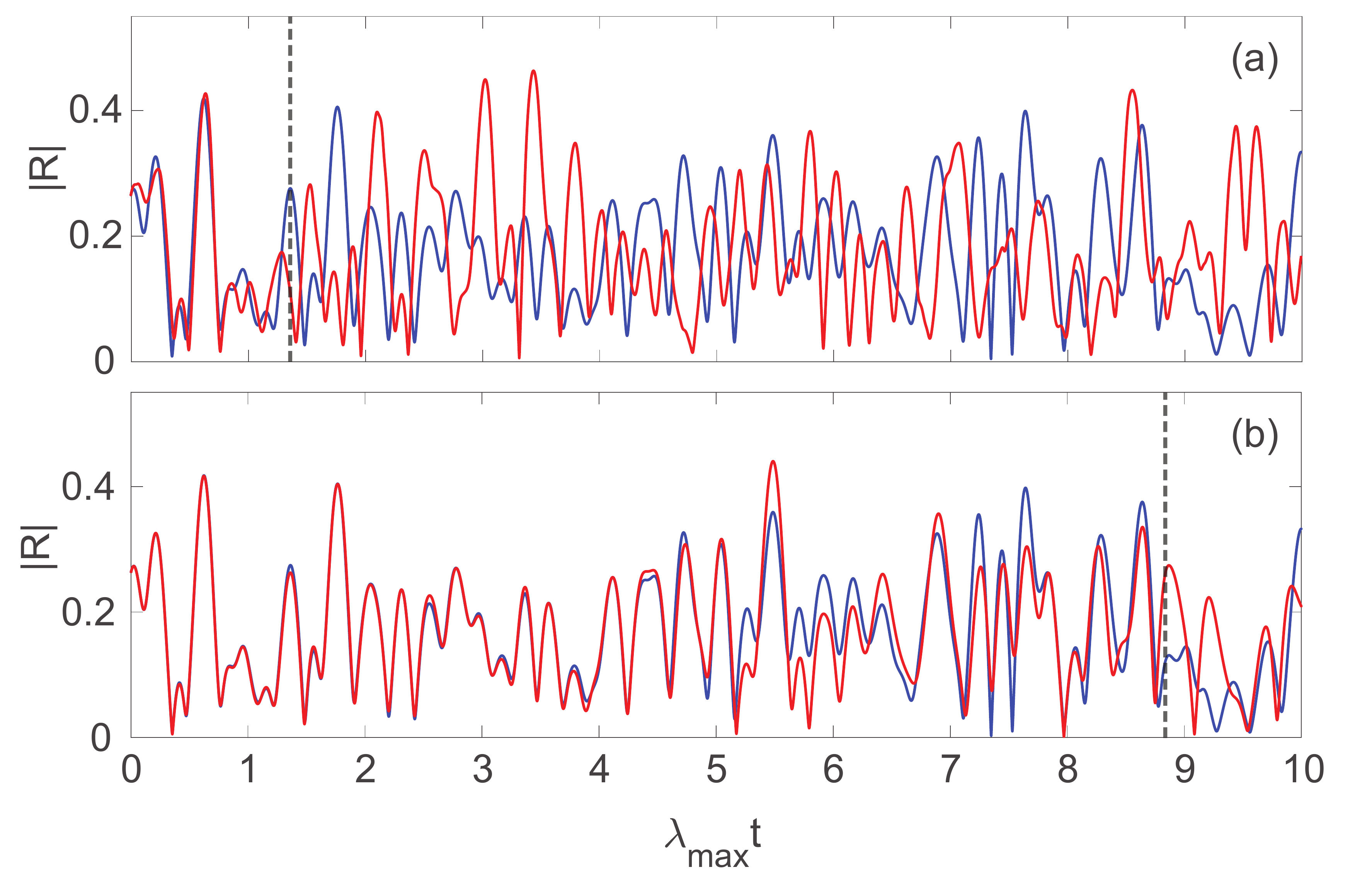}
\hfill
\caption{Parallel prediction of Kuramoto system with known neighbors. The blue curves are the data and red curves are predictions. The dotted lines in each plot denote the valid prediction time. (a) Using an all-to-all input matrix, i.e., all inputs map to all reservoir nodes. (b) Using an input matrix where each input is sent to a different disjoint set of reservoir nodes.}
\label{Suppfig:Win}
\end{figure}

We compare the performance of an all-to-all input matrix to one that is generated as mentioned above, with the input scaling, $\sigma$, optimized separately for each case ($\sigma=0.3$ for an all-to-all input matrix, for other hyperparameters see Table \ref{tab:hyperparams}). Supp. Fig. \ref{Suppfig:Win} clearly highlights the superior performance of the latter method for a Kuramoto oscillator network with all nodes having degree 3, where each reservoir in the parallel scheme (with known neighbors) receives 8 inputs from exactly 4 oscillators: 1 "assigned" + 3 neighbors. The assigned oscillator is the oscillator the reservoir is trained to predict. As we increase the number of inputs to a reservoir, it might be beneficial for each input to be connected to a non-exclusive subset of the reservoir. This would mean each input is still connected to a fixed number of reservoir nodes, but we can relax the criteria that each node in the reservoir is connected to exactly one input.\\

For parallel prediction with unknown neighbors, we must slightly alter our scheme. In this case, each of the input oscillators to a reservoir are not treated equally. Of the $N_{in}^{osc}$ input oscillators, we only know 1 ``true" input oscillator- the assigned oscillator that reservoir is intended to predict. We cannot distinguish the true neighbors from the false positives in the remaining $N_{in}^{osc}-1$ input oscillators. We can account for this by ``reserving" a larger part of the reservoir for the assigned oscillator and dividing the rest of the reservoir equally between all the inferred neighbors. This means, each of the 2 inputs from the assigned oscillator are connected to $N_{in}^{assign}/{2}$ reservoir nodes, while the remaining $N_{in}^{tot}-2$ inputs are each connected to $(N_r-N_{in}^{assign})/(N_{in}^{tot}-2)$ reservoir nodes. Here $N_r-N_{in}^{assign}$ is chosen to be an integer multiple of $N_{in}^{tot}-2$ and hence the number of reservoir nodes used in this case can only be set approximately equal to one with known neighbors. For example, in our system with known neighbors we set $N_r=200$. But for the case with unknown neighbors, where we set $N_{assign}=50$, if we have 7 inferred neighbors for a particular oscillator, the $N_r$ for the corresponding reservoir must be set to 204 to satisfy the criteria.

\section{RC hyperparameters}
We list the RC hyperparameters (Table \ref{tab:hyperparams}) used to plot the results shown in the paper. These hyperparameters were chosen by a subsequently iterative coarse grid search which approximately maximized the valid prediction time.
\begin{table}[H]
\def\arraystretch{1.5}
    \centering
        \begin{tabular}{|c||C|S|S|S|C|}
         \hline
         \ & Number of reservoir nodes & Spectral radius & Input scaling & Leak rate & Regularization parameter\\ 
         \ & $N_r$ & $\rho$ & $\sigma$ & $\alpha$ & $\beta$\\ 
         \hline
         Single Reservoir & 10000 & 0.9 & 0.1 & 0 & $10^{-7}$ \\ 
         \hline
         Parallel Scheme& &  & & & \\
         a) Known Neighbors & 200 & 0.9 & 0.6 & 0.1 & $10^{-9}$ \\ 
         b) Unknown Neighbors & $\approx 200$* & 0.9 & 0.5 & 0.1 & $10^{-8}$\\
         \hline
        \end{tabular}
    \caption{Values of reservoir hyperparameters obtained via a coarse grid search for a single reservoir prediction as well as for the parallel reservoir scheme. *Reservoir size set approximately to 200, for details see Section \ref{SuppSec2}.}
    \label{tab:hyperparams}
\end{table}
\newpage
\section{Node-level predictions}
\begin{figure}[h]
\centering
\includegraphics[width=.85\textwidth]{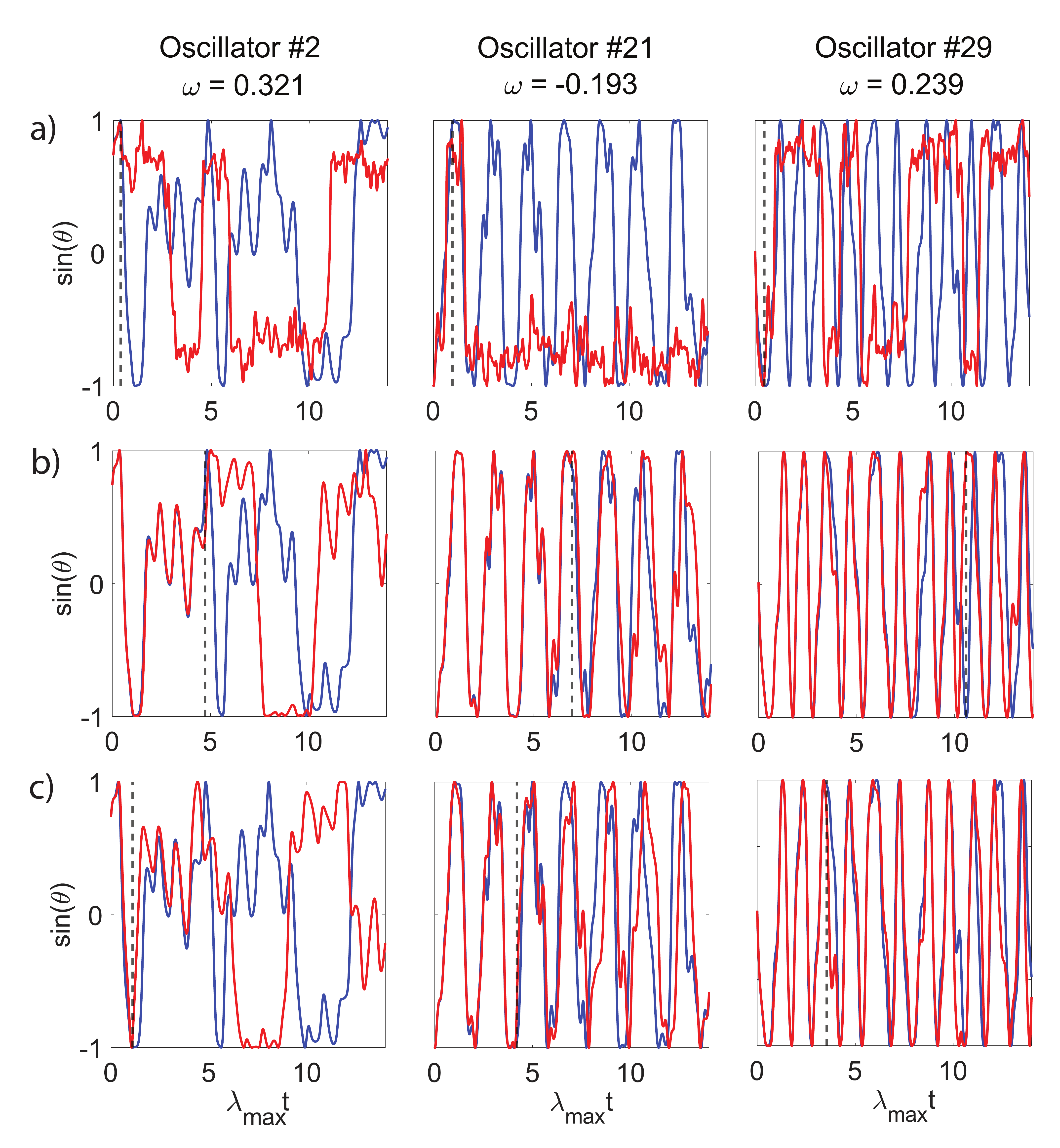}
\hfill
\caption{Prediction of the dynamics of three representative individual oscillators. The natural frequency of the oscillators is noted above. The blue curves are the data and red curves are predictions. The vertical dotted lines in each plot denote the valid prediction time. (a) Single, non-parallel RC prediction using a large reservoir ($N_r=10000$), (b) Parallel prediction (with known network links) using 50 separate reservoirs each having modest size ($N_r=200$), (c) Parallel prediction (with unknown network links) using 50 separate reservoirs each having modest size ($N_r\approx200$). For row (c) the network structure is estimated by using transfer entropy as a metric for inferring network edges.}
\label{Suppfig:nodeLevel}
\end{figure}

While we study the predictive performance of our ML scheme, it is also important to note the node-level details. Supp. Fig. \hyperref[Suppfig:nodeLevel]{3(a)}, shows that a single large reservoir ($N_r=10000$) is unable to capture the dynamics of individual oscillators in the network. However, our parallel schemes with known edges [Supp. Fig. \hyperref[Suppfig:nodeLevel]{3(b)}], as well as unknown edges [Supp. Fig. \hyperref[Suppfig:nodeLevel]{3(c)}], are able to capture the dynamics of the individual oscillators. The parallel schemes, also give good predictions of the system ``climate" (i.e. the prediction still seems to resemble the original dynamics) even after the marked valid prediction time (vertical dotted lines in Supp. Fig. \hyperref[Suppfig:nodeLevel]{3}). 
